\documentclass{article}

\usepackage{microtype}
\usepackage{graphicx}
\usepackage{subfigure}
\usepackage{booktabs} 

\usepackage{hyperref}

\usepackage[accepted]{sysml2019}

\usepackage{xcolor}
\usepackage{multirow}
\usepackage[normalem]{ulem}
\usepackage{fancyhdr}
\usepackage{amsmath}
\usepackage{xspace}
\usepackage{soul}
\usepackage{placeins}

\usepackage{url}
\usepackage{etex}
\usepackage{etoolbox}
\usepackage{nicefrac}
\usepackage{changepage}
\usepackage{array,booktabs,multirow}
\usepackage{afterpage}
\usepackage{pbox}
\usepackage{paralist}
\usepackage{mdwlist}
\usepackage{hhline}
\usepackage{lipsum} 
\usepackage{adjustbox}
\usepackage{tabu}
\usepackage{makecell}
\usepackage{color, colortbl}
\usepackage{amsmath,amssymb,latexsym}
\usepackage{enumitem}

\usepackage{amsfonts}

\usepackage{pifont}

\usepackage{threeparttable}
\usepackage{xpatch}
\usepackage[bottom]{footmisc}

\makeatletter
\chardef\TPT@@@asteriskcatcode=\catcode`*
\catcode`*=11
\xpatchcmd{\threeparttable}
  {\TPT@hookin{tabular}}
  {\TPT@hookin{tabular}\TPT@hookin{tabu}}
  {}{}
\catcode`*=\TPT@@@asteriskcatcode
\makeatother

\makeatletter                                                                             
\newcommand{\thickhline}{%
    \noalign {\ifnum 0=`}\fi \hrule height 1.2pt                                          
    \futurelet \reserved@a \@xhline                                                       
}                                                                                         
\makeatletter                                                                             
\newcommand{\midhline}{%
    \noalign {\ifnum 0=`}\fi \hrule height 1pt                                            
    \futurelet \reserved@a \@xhline                                                       
}    

\makeatletter
\def\SOUL@hlpreamble{%
    \setul{0ex}{2ex}
    \let\SOUL@stcolor\SOUL@hlcolor
    \SOUL@stpreamble
}

\makeatletter  
\def\@eqnnum{{\normalfont\normalcolor[\theequation]}}  
\makeatother

\newcommand{\ceil}[1]{\lceil #1 \rceil}

\newcommand{\abbrfnt}{}
\newcommand{\abbrev}[1]{{\mbox{\abbrfnt{#1}}}\xspace}

\newcommand{\eq}[1]{Eq.~\ref{#1}\xspace}

\makeatletter
\renewcommand{\fnum@table}{Tab. \thetable}
\makeatother
\newcommand{\tab}[1]{Tab.~\ref{#1}\xspace}

\makeatletter
\renewcommand{\fnum@figure}{Fig. \thefigure}
\makeatother
\newcommand{\fig}[2][]{Fig.~\ref{#2}#1\xspace}

\newcommand{\mbs}{\abbrev{MBS}}

\newcommand{\wavecore}{\abbrev{WaveCore}}
\newcommand{\scratch}[1]{}

\newcommand{\mdel}[2]{}

\newcommand{\mattancut}[2]{}

\newcommand{\FIXME}[1]{}

\sysmltitlerunning{Mini-batch Serialization: CNN Training with Inter-layer Data Reuse}

\begin{document}

\twocolumn[
\sysmltitle{Mini-batch Serialization: \\ CNN Training with Inter-layer Data Reuse}

\sysmlsetsymbol{equal}{*}

\begin{sysmlauthorlist}
\sysmlauthor{Sangkug Lym}{ut}
\sysmlauthor{Armand Behroozi}{mi}
\sysmlauthor{Wei Wen}{du}
\sysmlauthor{Ge Li}{ut}
\sysmlauthor{Yongkee Kwon}{ut}
\sysmlauthor{Mattan Erez}{ut}
\end{sysmlauthorlist}

\sysmlaffiliation{ut}{The University of Texas at Austin}
\sysmlaffiliation{mi}{University of Michigan}
\sysmlaffiliation{du}{Duke University}

\sysmlcorrespondingauthor{Sangkug Lym}{sklym@utexas.edu}
\vskip 0.3in

\begin{abstract}
Training convolutional neural networks (CNNs) requires intense computations and high memory bandwidth. We find that bandwidth today is over-provisioned because most memory accesses in CNN training can be eliminated by rearranging computation to better utilize on-chip buffers and avoid traffic resulting from large per-layer memory footprints. We introduce the \mbs CNN training approach that significantly reduces memory traffic by partially serializing mini-batch processing across groups of layers. This optimizes reuse within on-chip buffers and balances both intra-layer and inter-layer reuse. We also introduce the \emph{WaveCore} CNN training accelerator that effectively trains CNNs in the \mbs approach with high functional-unit utilization. Combined, WaveCore and \mbs reduce DRAM traffic by 75\%, improve performance by 53\%, and save 26\% system energy for modern deep CNN training compared to conventional training mechanisms and accelerators.
\end{abstract}


]
\printAffiliationsAndNotice{} 

\section{introduction}
\label{sec:intro}
Convolutional neural networks (CNNs) are the state of the art for various vision applications. Training CNNs requires hundreds of thousands of compute- and data-intensive iterations.
We observe that CNN training on current systems requires 3--4 times more off-chip memory bandwidth than necessary, reducing performance and wasting energy.
We present a new training mechanism that significantly reduces bandwidth demands for the same arithmetic performance by better exploiting locality. We then develop a complementary accelerator that dramatically lowers training cost and time.

\begin{figure}[t!]
    \centering
    \includegraphics[width=0.46\textwidth]{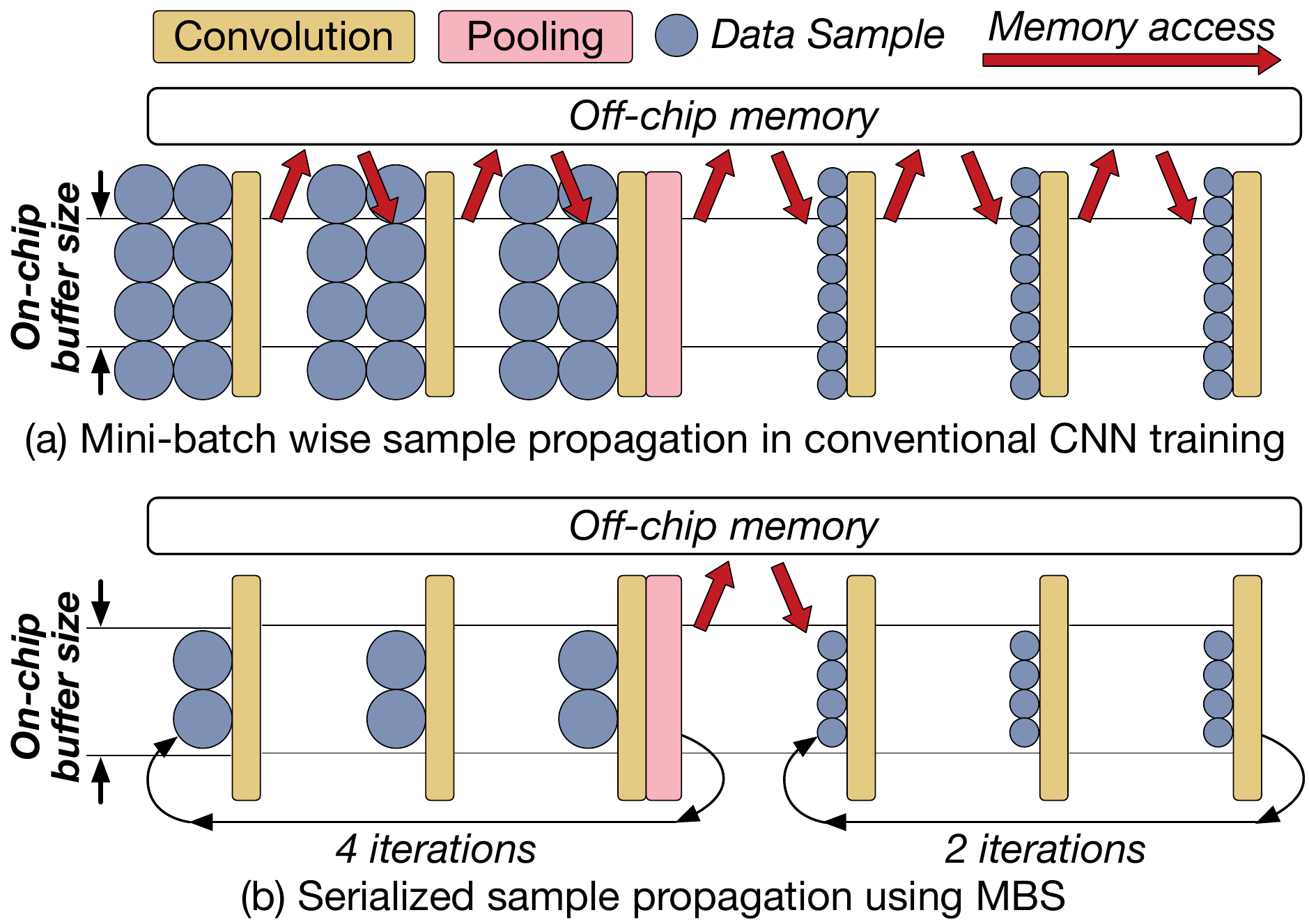}
    \vspace*{-3mm}
    \caption{A toy CNN architecture. \mbs restricts per-layer memory footprint size smaller than the on-chip buffer.}
    \label{fig:main}
  \vspace*{-4mm}
\end{figure}

Conventional CNN training propagates data in lockstep across network layers for an entire mini-batch (typically  32--512 samples per processor~\cite{szegedy2015going,szegedy2017inception,he2016deep}). Large mini-batches have per-layer memory footprints that exceed typical on-chip buffer capacity, resulting in high off-chip memory traffic (\fig{fig:main}{a}). Directly applying locality techniques used in CNN inference~\cite{parashar2017scnn,alwani2016fused} to training is ineffective because such techniques do not optimize locality across large mini-batches, and their design is not compatible with feature normalization~\cite{ioffe2015batch}.

Our \emph{mini-batch serialization} (\mbs) approach reduces memory traffic specifically for CNN training and exploits data reuse across layers (inter-layer data)---a first for training and for modern networks that include multi-branch modules and normalization layers. As illustrated in \fig{fig:main}{b}, \mbs breaks a mini-batch into \emph{sub-batches} to reduce the per-layer memory footprint such that the inter-layer data of an entire sub-batch fits in on-chip buffers. \mbs uses a different number of samples per sub-batch (sub-batch size) for different layers because down-sampling layers decrease the size of each feature and, hence, the total volume of features.

\mbs forms groups of layers such that each group has the same sub-batch size. Each sub-batch is then propagated with most inter-layer data staying on chip across the layers of a group; data needed during back propagation is stored off chip as well. Otherwise, layer output data is only written and later read from main memory between groups. \mbs optimizes sub-batch sizes and layer grouping to balance data reuse between layers with reuse of parameters (weights) within a layer---weights are re-read for every sub-batch. \mbs cuts memory traffic by $4.0\times$ compared to conventional layer-by-layer mini-batch training across a number of popular deep CNNs.

While \mbs optimizes locality and reduces memory traffic, applying \mbs to modern deep CNNs and using it in a training accelerator introduces two challenges. The first is normalization as sub-batches prohibit the use of batch normalization. We therefore adapt group normalization~\cite{wu2018group} for use in the \mbs flow and demonstrate the effectiveness of \mbs training.

The second challenge is that \mbs reduces per-layer parallelism, which potentially lowers the utilization of arithmetic units in the accelerator. We address this issue in two ways. First, we use the \emph{im2col} (image-to-column) convolution algorithm that is commonly used by GPUs. It casts a convolution operation as a general matrix multiply (GEMM)~\cite{chetlur2014cudnn}. With im2col, the reduced parallelism from small sub-batches is effectively compensated for by the size of other feature dimensions. Second, we modify a traditional systolic array processing core, as used by some commercial accelerators~\cite{jouppi2017datacenter,lu2017flexflow}, to better execute the tall and skinny GEMMs needed for im2col.

GEMM is blocked to utilize a systolic array, but idle time exists between the execution of any two blocks in a conventional design. To avoid this idle time, we augment each processing element with one additional 16b register that double buffers inputs and eliminates gaps between blocks. Our \emph{\wavecore} accelerator with \mbs achieves compute-unit utilization that is within $3\%$ that of conventional training.

We use three recent deep CNNs to evaluate \mbs and \wavecore: ResNet \cite{he2016deep}, Inception v3 \cite{szegedy2015going}, and Inception v4 \cite{szegedy2017inception}. We show that \mbs saves DRAM accesses by 78\% 71\%, 74\%, improves training performance by 66\%, 36\%, 40\%, and saves 30\%, 24\%, and 24\% energy for ResNet50 and Inception v3 and v4, respectively. We also demonstrate that \mbs enables high-performance training accelerators that use much more affordable but slower off-chip memory (e.g., LPDDR)---even with 60\% less memory bandwidth, training performance is still 24\% above the baseline design.

We summarize our contributions below:
\begin{itemize}[topsep=-6pt,parsep=-6pt,partopsep=0pt,leftmargin=10pt,labelwidth=6pt,labelsep=4pt]
\item
    We introduce Mini-Batch Serialization (\mbs), a hardware-resource aware CNN training optimization that significantly reduces off-chip memory traffic and can thus accelerate CNN training and reduce training cost. \mbs balances intra- and inter-layer locality and cuts DRAM traffic by $4.0\times$ for modern CNNs.
\item 
    We show how \mbs exploits locality within multi-branch modules to achieve the $4.0\times$ traffic reduction (traffic increases by 20\% without this multi-branch optimization).
\item 
    We augment a conventional systolic array architecture and optimize it to effectively accelerate {\mbs}-based training. Our WaveCore accelerator maintains high processing-element utilization for modern CNNs and utilizes \mbs to provide high performance with both high-bandwidth HBM2 DRAM (as used by GPUs and Google's TPU v2 and v3) and even lower-cost and higher-capacity GDDR5 and LPDDR4 DRAM systems.
\end{itemize}


\section{Data Locality in CNN Training}
\label{sec:inter-layer}

\begin{figure}[t]
    \centering
    \includegraphics[width=0.46\textwidth]{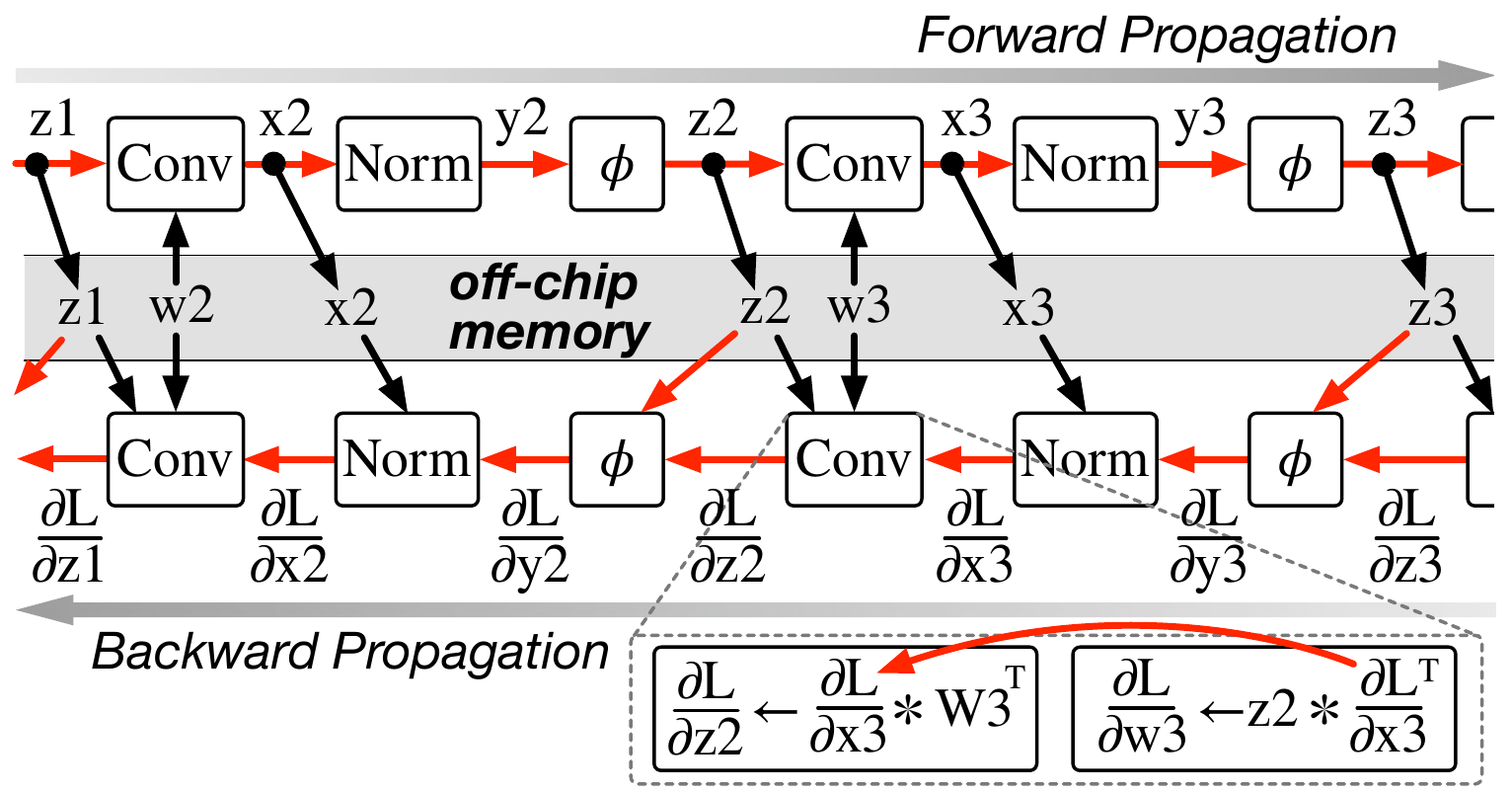}
    \vspace*{-2mm}
    \caption{Dataflow in forward and backward propagations. Red arrows show the reusable data between layers.}
    \label{fig:train}
  \vspace*{-4mm}
\end{figure}

CNN training consists of forward and back propagation phases. \fig{fig:train} illustrates the major data elements needed for training and their reuse patterns with red arrows indicating opportunities for on-chip buffers to reduce memory bandwidth requirements and black arrows indicating accesses to main memory. In both phases there is direct producer-consumer locality between layers---\emph{inter-layer data} that can be buffered if it is not too large. The outputs of convolution, normalization, and activation layers in forward propagation (\(x\), \(y\), and \(z\) in the figure) are immediately used by their following layers. Normalization layers exhibit additional reuse because they iterate over inputs to first compute the mean and variance before normalizing the data~\cite{ioffe2015batch}. The convolution outputs and the activations are stored in off-chip memory for reuse in back propagation, because their large storage requirements and long data reuse distance prevent on-chip buffering.

Back propagation exhibits even greater potential for inter-layer reuse. The loss gradients (with respect to \(x\)) are reused twice by a convolution layer to compute the gradients of weights and loss (with respect to \(z\)). Also, the convolution output stored in memory is reused multiple times to compute the gradients of the normalization layer parameters and the loss gradients (with respect to \(x\)). Activations read from memory are also used twice: \(z\) is used for convolution gradients and the derivative of \(z\) for activation gradients.

\textbf{The Problem with CNN Training Memory Footprint.}
CNN training is typically done with mini-batches of 32--512 samples (possibly distributed across multiple processors)~\cite{sutskever2013importance}. Larger mini-batches reduce model parameter update frequency and training iterations, thus reducing training time and energy~\cite{li2014efficient,goyal2017accurate}. An additional benefit is that larger data parallelism can be used to maintain high compute unit utilization and help in distributing work across multiple processors~\cite{das2016distributed}.

\begin{figure}[t]
    \centering
    \includegraphics[width=0.48\textwidth]{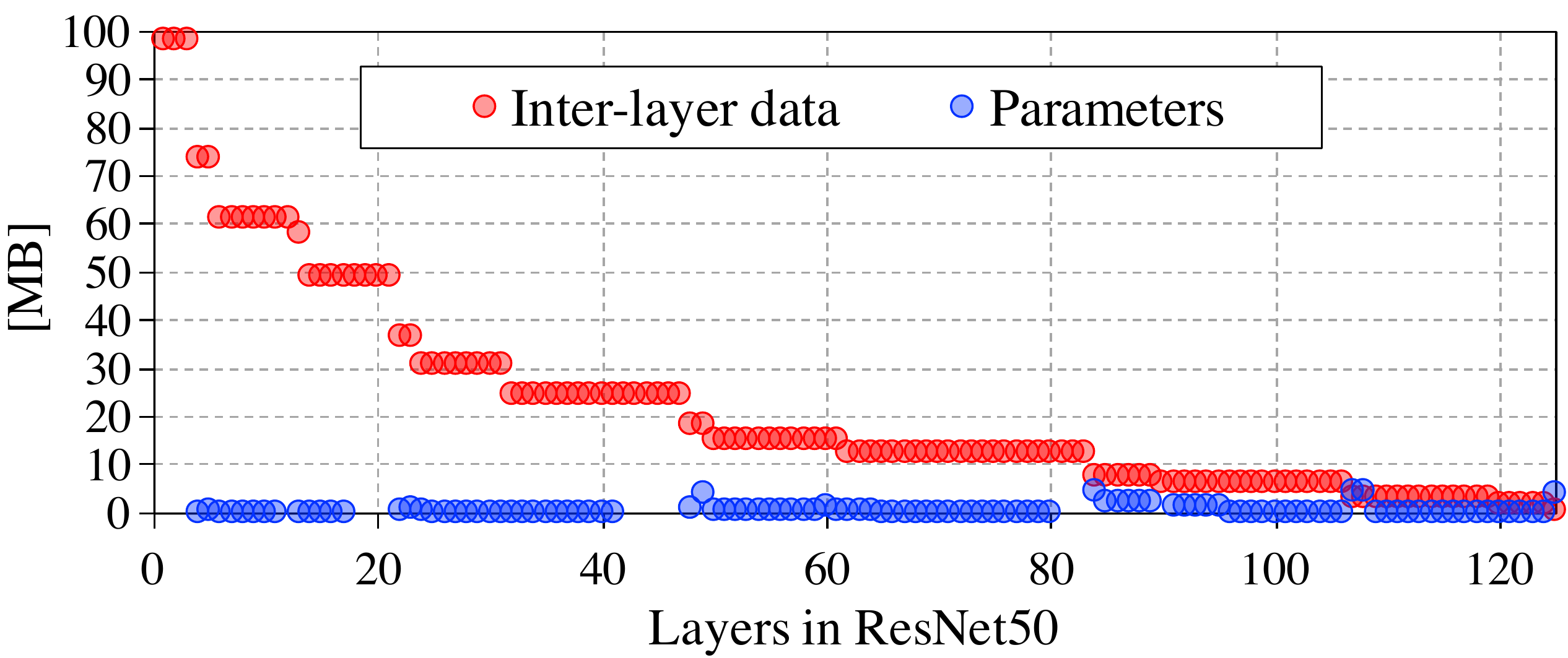}
    \vspace*{-8mm}
    \caption{The size of inter-layer data and parameters of each layer in ResNet50 (sorted by inter-layer data size).}
    \label{fig:footprint}
  \vspace*{-5mm}
\end{figure}

However, a larger per-processor mini-batch with many samples increases the memory footprint of each layer, limiting the opportunity to reuse data on chip. \fig{fig:footprint} shows the per-layer footprint of ResNet50 with a mini-batch size of 32 and a word size of 16b (in the forward phase). Only 9.3\% of inter-layer data can be reused even with 10MiB on-chip storage, leading to very significant memory bandwidth waste for storing and refetching data. This problem is even more severe for larger mini-batch sizes, which are desirable as per-processor arithmetic performance and main memory capacity improve.

\section{Mini-Batch Serialization}
\label{sec:mbs}
The primary goal of \mbs is to improve reuse by exploiting inter-layer data locality. The key to \mbs is partially serializing a mini-batch (propagating a small sub-set of a mini-batch at a time) to control per-layer memory footprint without impacting training accuracy. \mbs is based on our insight that if the data synchronization points for functional correctness are maintained and an appropriate normalization algorithm is adapted, even processing a single sample at a time through all network layers does not alter the training result. The trivial serialization of one sample at a time, however, has two crucial drawbacks.

First, while baseline training reads weights and writes weight gradients just once per layer, full serialization re-reads weights and partial gradient sums for each sample and updates the partial sums once per sample as well. Second, data parallelism within a single sample can be limited in some layers, degrading resource utilization and performance
(especially when mapping to a highly-efficient systolic architecture). 

An improvement on full serialization is to process multiple samples at a time (a \emph{sub-batch}) to provide some intra-layer weight reuse and extra parallelism, as long as the footprint at any point in the sub-batch does not exceed the on-chip buffer capacity. The entire mini-batch is then processed in several sub-batch iterations. However, the footprints of early layers are large and only a small sub-batch can be formed (1--2 samples), limiting the benefits of this approach.

\textit{\mbs goes much further and balances locality of intra-layer weight reuse and parallelism with inter-layer locality.} We do this by varying the number of samples per sub-batch across layers such that layers that can support more samples require fewer iterations and can benefit from the greater parallelism and locality. This is possible because down-sampling (pooling and strided convolution) layers decrease feature size and volume for deeper layers.

\begin{figure}[t!]
    \centering
    \includegraphics[width=0.48\textwidth]{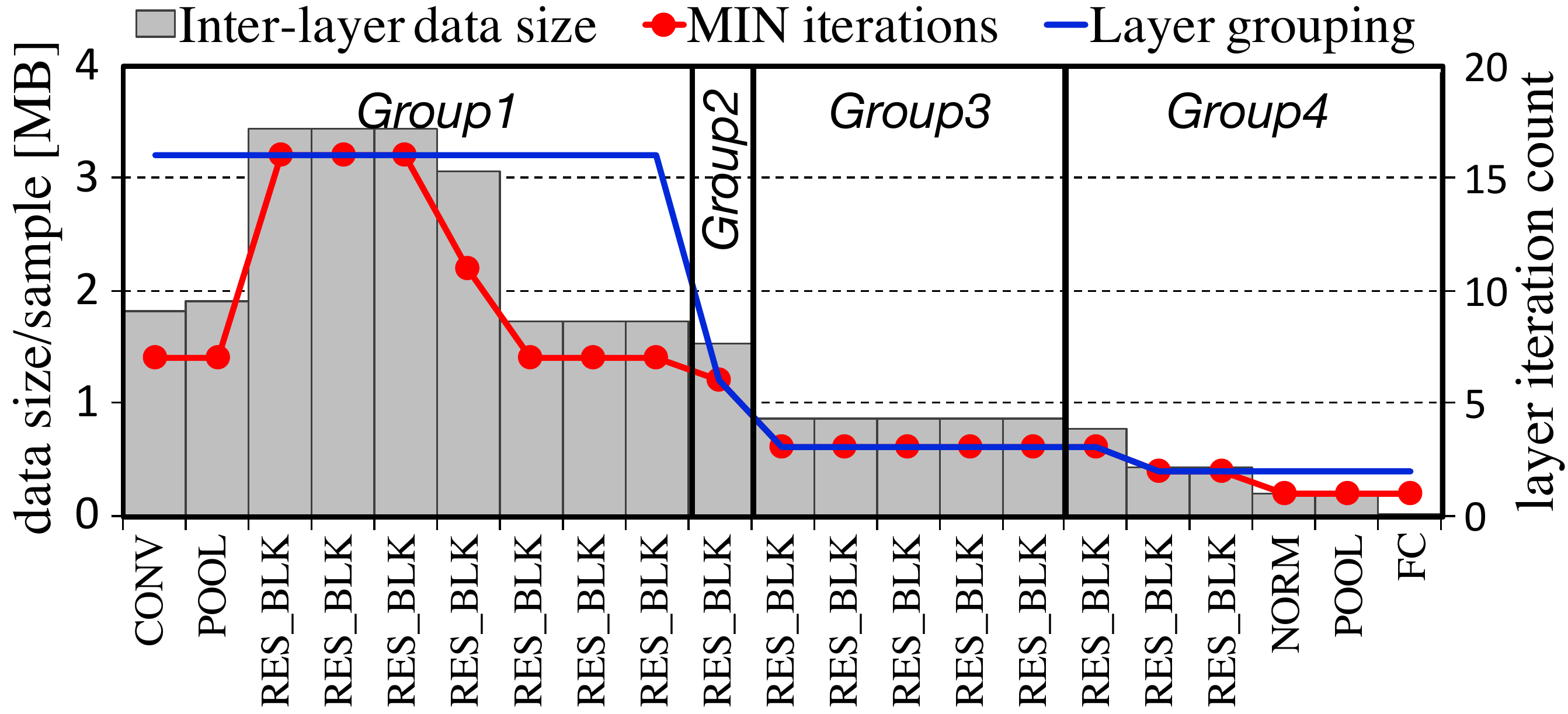}
    \vspace*{-7mm}
    \caption{Per-block inter-layer data size, required layer iterations, and \mbs layer grouping for ResNet50 with 32 samples.}
    \label{fig:mbs1}
  \vspace*{-1mm}
\end{figure}

\begin{figure}[t!]
    \centering
    \includegraphics[width=0.47\textwidth]{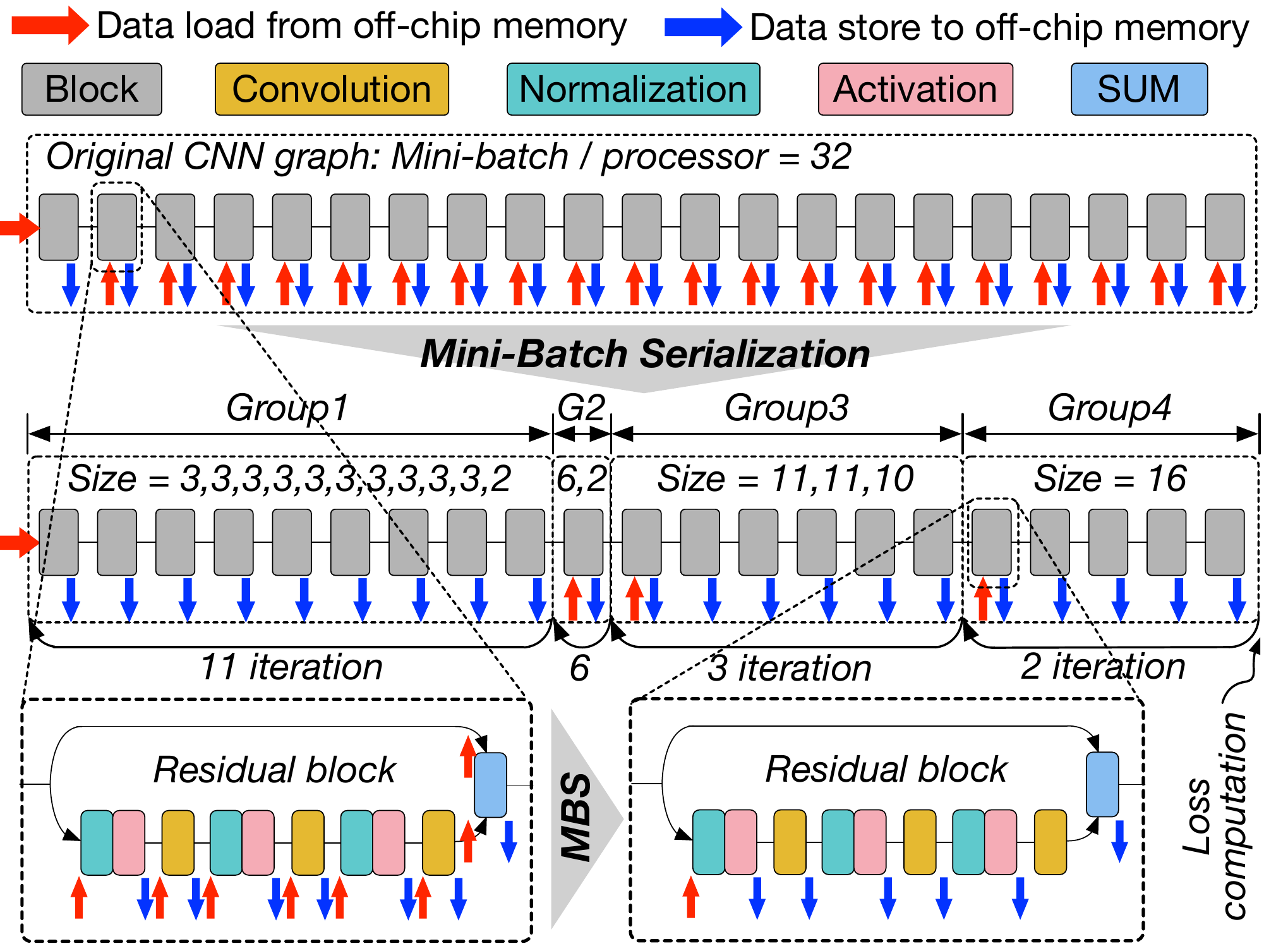}
    \vspace*{-2mm}
    \caption{Baseline and \mbs ResNet training flow.}
    \label{fig:mbs}
  \vspace*{-5mm}
\end{figure}

\begin{figure*}[t]
\centering
\begin{minipage}{.42\linewidth}
    \includegraphics[width=\textwidth]{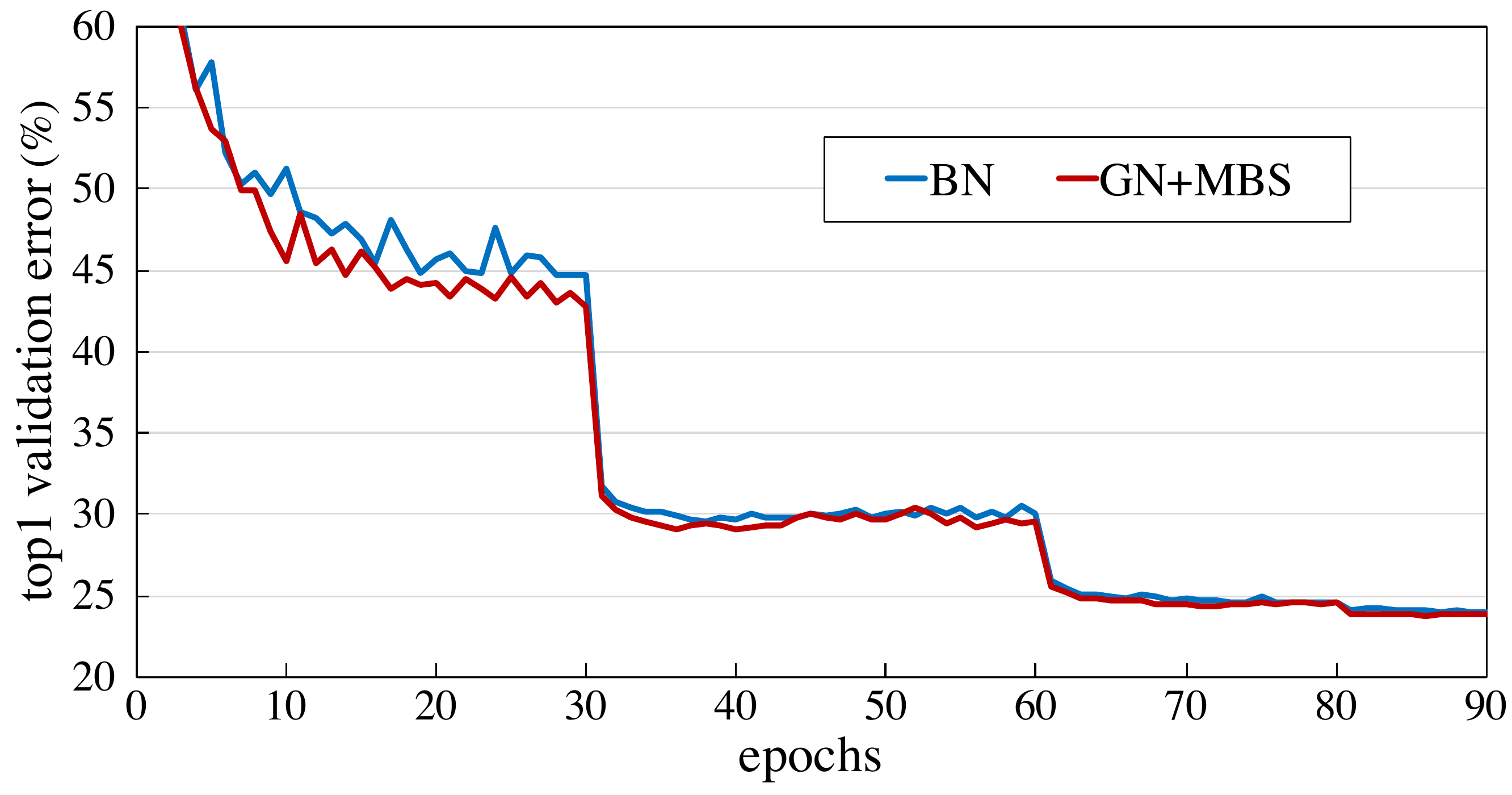}
\end{minipage}
\begin{minipage}{.56\linewidth}
    \includegraphics[width=\textwidth]{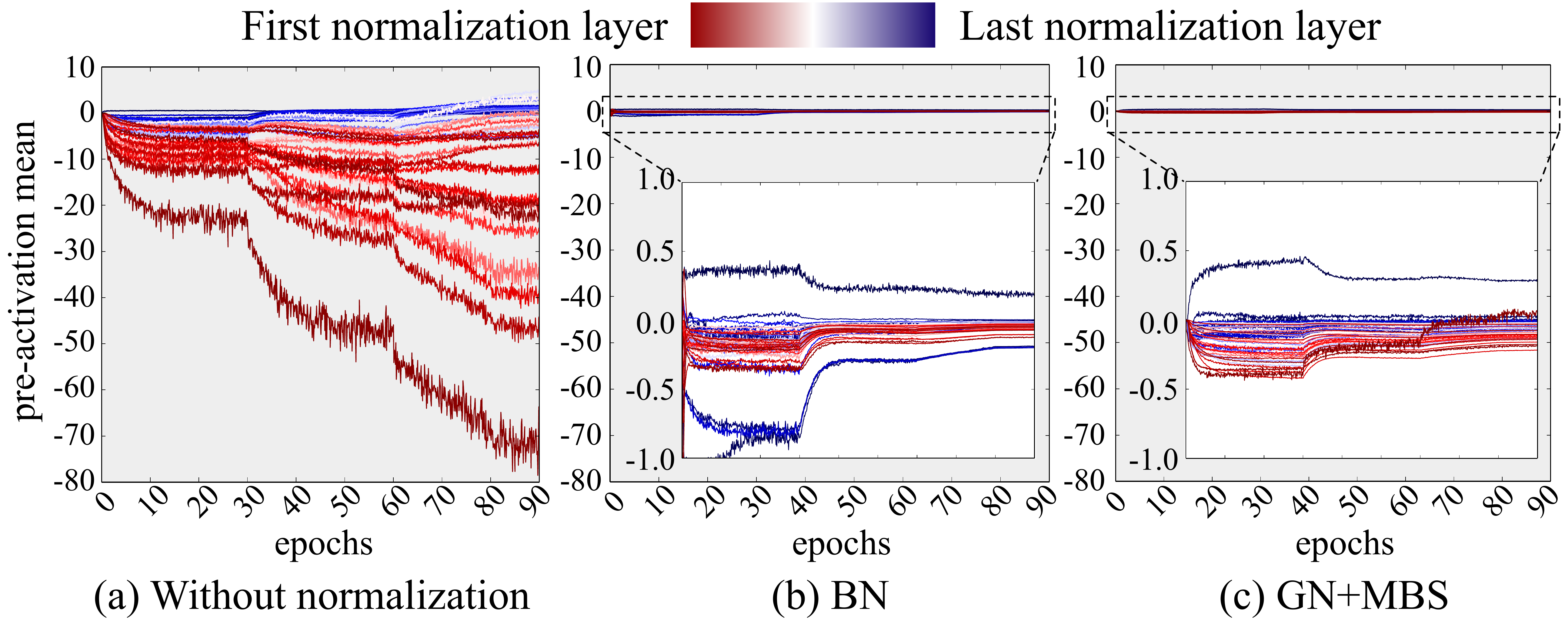}
\end{minipage}
\vspace*{-1mm}
\caption{ResNet50 training with GN + MBS and BN: validation error (left) and pre-activation mean of each normalization layer with BN and GN zoomed (right). Mini-batch size of 128 distributed across 4 GPUs, an initial learning rate 0.05~\cite{bottou2018optimization}, and learning rate decays of 0.1 at epochs 30, 60, and 80. \scriptsize{(The code is available at \url{https://bitbucket.org/lph\_tools/mini-batch-serialization}.)}}
\vspace*{-3mm}
\label{fig:sgn_train}
\end{figure*}

\textbf{Layer Grouping Optimizes Reuse.}
Optimizing layer groups balances intra- and inter-layer locality tradeoffs. The \mbs algorithm forms initial layer groups by grouping adjacent layers that require the same number of sub-batch iterations. This is shown in \fig{fig:mbs1} where grey vertical bars represent the data volume required for the inter-layer data per layer (or one multi-branch module {\bf\emph{block}}) of ResNet50, and the red line represents the resulting minimal sub-batch iteration count for each layer. Then, layer groups are merged to improve overall locality: groups are merged by reducing the sub-batch size of one group to that of an adjacent group. The first group then requires more iterations (with more weight and gradient accesses), but inter-layer reuse increases across the two layers where the groups meet. The resulting grouping for this optimization for ResNet50 is shown with the blue line in \fig{fig:mbs1}.\footnote{We also experimented with an optimal grouping of layers using exhaustive search, which improved traffic and performance by roughly $1\%$ compared to our greedy optimization.}

The mini-batch is then processed in several sub-batch iterations (\(\ceil{\frac{\mathit{mini-batch\ size}}{\mathit{sub-batch\ size}}}\)) within each group as shown in \fig{fig:mbs}, which emphasizes how locality is increased and memory traffic reduced across features and weights.

\textbf{Data Reuse Within Multi-Branch Modules.}
\fig{fig:mbs} also shows how \mbs applies the same sub-batch approach to a multi-branch residual module of ResNet50. Such multi-branch modules are common in CNN architectures and offer additional reuse opportunities. Both the residual and shortcut branches share an input, and when they merge, their outputs are summed. Therefore, the module inputs should stay on chip until both paths have consumed them, and the output of the shortcut branch should stay on-chip while the main path output is computed. \mbs does this by provisioning buffer space based on the needs of multi-branch \emph{blocks}, where a block includes all the branches that share split and merge points---\mbs essentially treats such a block as a layer for optimizing locality.

Maintaining locality for such shared nodes leads to additional storage requirements. The per-sample size is calculated by \eq{eq:storage_residual} where: \(D_{in}\) and \(D_{out}\) indicate the sizes of the main-branch input and output; \(D_{shortcut}\) is the size of the shortcut path output;  \(L\) is the number of layers in the main branch; and \(b\) and \(l\) represent a specific branch and layer.
\begin{equation}
\label{eq:storage_residual}
\scriptsize
\begin{split}
&\frac{Space}{Sample} = \! \max\limits_{1 \leq b \leq 2,\ 1 \leq l \leq L} \! D_{in}(b,l) + D_{out}(b,l) + D_{cond}(b,l)\\
&D_{cond}(b,l) = (b \!=\! 1 \ \&\  l \!\neq\! 1)D_{block\_in} + (b \!\neq\! 1)D_{block\_out}
\end{split}
\end{equation}
Similarly, for inception modules~\cite{szegedy2015going,szegedy2017inception}, the block input is reused between branches, and the concatenated block output is eventually reused in the following layer. Therefore, \mbs keeps both the block input and output on chip while executing the branches. The space required is shown in \eq{eq:storage_inception}, where \(B\) indicates the number of branches in a module and other notation is as above.
\begin{equation}
\label{eq:storage_inception}
\scriptsize
\begin{split}
&\frac{Space}{Sample}  \!=\! \max\limits_{1 \leq b \leq B,\ \!1 \leq l \leq L}\!\!D_{in}(b,l) \!+\! D_{out}(b,l) \!+\! D_{cond}(l) \\
&D_{cond}(l) \!= (l \!\neq\! 1)D_{block\_in} + (l \!\neq\! L)D_{block\_out} 
\end{split}
\end{equation}

\textbf{Back Propagation.}
In back propagation, \mbs optimizes locality for both newly computed results and for data reloaded from the forward path. For example, as shown in \fig{fig:train}, \mbs reuses the reloaded gradients more than once. Furthermore, both convolution and ReLU layers use activations from the forward path. However, only the gradient of ReLU is needed, which is always exactly 0 (for negative activations) and exactly 1 (for positive); thus, MBS uses a single bit per ReLU gradient instead of a 16b number. We also allocate buffer space for normalization layers to reuse their inputs to compute their gradient and loss. As in the forward pass, reuse in back propagation is made possible by \mbs processing one sub-batch at a time. 

\textbf{Data Synchronization.}
\mbs maintains the original synchronization points across the entire mini-batch. Therefore, \mbs accumulates the partial gradients of all learning parameters across all sub-batches. 
This requires storing partial results to memory, which is not needed in the conventional flow. 
However, this overhead is dwarfed by the improved reuse of layer outputs, especially considering that deeper layers with large weights are iterated over only a few times.


\subsection{Feature Normalization in \mbs}
\label{subsec:sgn}

While batch normalization (BN) is widely used in many modern CNNs, it is incompatible with \mbs because BN requires many samples to work well and improve accuracy~\cite{ioffe2015batch}---\mbs cannot serialize computation if data across an entire mini-batch (per processor) is needed for normalization. Instead of using BN, we adapt group normalization (GN)~\cite{wu2018group} to \mbs.
GN normalizes across features within a subset of channels in a single sample, as opposed to across an entire per-processor mini-batch. Thus, GN can be made compatible with \mbs.

To use GN with \mbs, the per-channel GN scale and shift parameters must be re-fetched at every sub-batch iteration within a layer group. Additionally in backpropagation, the gradients of these parameters must be accumulated across all sub-batches just like the weights of convolution layers. However, since the size of these parameters is only two times the number of channels per layer, they can easily be stored in the on-chip buffer and incur no overhead.

We confirm previous results and demonstrate that both GN and BN provide comparable training effectiveness (validation accuracy of 76.0\% and 76.2\% for BN and GN\(+\)MBS, respectively). \fig{fig:sgn_train} compares the validation error curves with BN and {\mbs}-GN when training ResNet50 on ImageNet~\cite{imagenet_cvpr09}. \fig{fig:sgn_train} also shows that both {\mbs}-GN and BN provide similar normalization, in that both have similar pre-activation (output of normalization) distributions across layers (unlike training without normalization).


\section{WaveCore Accelerator}
\label{sec:arch}
We explain the WaveCore accelerator operation in two parts. First, we discuss the core compute engine of WaveCore and any modifications we make to adapt the conventional systolic array for \mbs. Second, we describe the overall accelerator architecture, which includes additional components for processing normalization, activation, and pooling layers. We also estimate the area and power of WaveCore.

\subsection{Systolic Array Core}
\label{sub_sec:systolic}
WaveCore uses a large systolic array as its main compute unit. A systolic array is (typically) a two-dimensional mesh of many simple and efficient processing elements (PEs). At each cycle of a kernel, each PE applies the same computation to its inputs and then passes the computed result or its unmodified inputs to one or more of its neighbors. All PEs communicate only with adjacent PEs such that there is minimal data movement and high computational concurrency~\cite{kung1982systolic}. Computation consists of pipelining inputs from the top and left (for example) edges of the array and obtaining results at the bottom. The large compute throughput required for convolutional and fully-connected layers, along with the repetitive computation and large data reuse are a good match for a systolic array, as found in Google's TPU ML accelerators~\cite{jouppi2017datacenter,tpuv2}. Like prior work, our proposed PE has mixed precision units: 16b inputs are multiplied with accumulation performed in 32 bits to reduce both computation and data traffic overheads~\cite{micikevicius2017mixed}. Also like prior work~\cite{tpuv2}, we use a $128 \!\times\! 128$ systolic array for high performance and to circumvent power delivery challenges.

\mattancut{Criticizing TPU is not particularly helpful here. I think we just get to the point and talk about what we do.}{Depending on the data reuse pattern among PEs, various systolic dataflows (data movement between PEs) can be needed.
One popular method described in TPU~\cite{jouppi2017datacenter} uses 2D-systolic array architecture where both weights and activation flow to the PE array for multiplication and accumulation in the column and row directions respectively.
However, based on our analysis, directly mapping convolution layer workload to systolic dataflow, like in the TPU, degrades the compute unit utilization when the mini-batch size is small.}

A systolic computation is often divided into multiple \emph{waves}, where each wave proceeds with inputs flowing toward outputs without any stalls or changes to the computational pattern. Between waves, it is sometimes necessary to let the pipeline through the array drain and then refill. This introduces idle time which reduces utilization and hence hurts performance and efficiency. Convolution and matrix operations have efficient systolic implementations that have little idle time if an entire mini-batch is processed together. However, \mbs processes an often small sub-batch, which significantly reduces the utilization and performance of a conventional systolic array design. We address this challenge and maintain high systolic array utilization for \mbs using a combination of two techniques. 

\textbf{Maintaining High Compute Unit Utilization with im2col.}
First, instead of directly mapping a convolution computation to a systolic array, we use a method that transforms a convolution into a matrix multiplication. We do this because efficient direct convolution on a systolic array requires tuning for every possible sub-batch size, which is difficult to do with the \mbs approach which optimizes groupings to arbitrary size. We use the \emph{im2col} (image-to-column) general matrix-matrix multiplication (GEMM) algorithm for convolution, which is commonly used in GPU-accelerated kernels~\cite{chetlur2014cudnn}. Convolution with \emph{im2col} rearranges the address pointers to the convolution inputs in a way that is straightforward to feed into a systolic array. The GEMM dimensions (\(G_h, G_w, K\)) are determined by the convolution configurations as summarized in \tab{tab:im2col}.

\begin{figure}[t]
    \centering
    \includegraphics[width=0.47\textwidth]{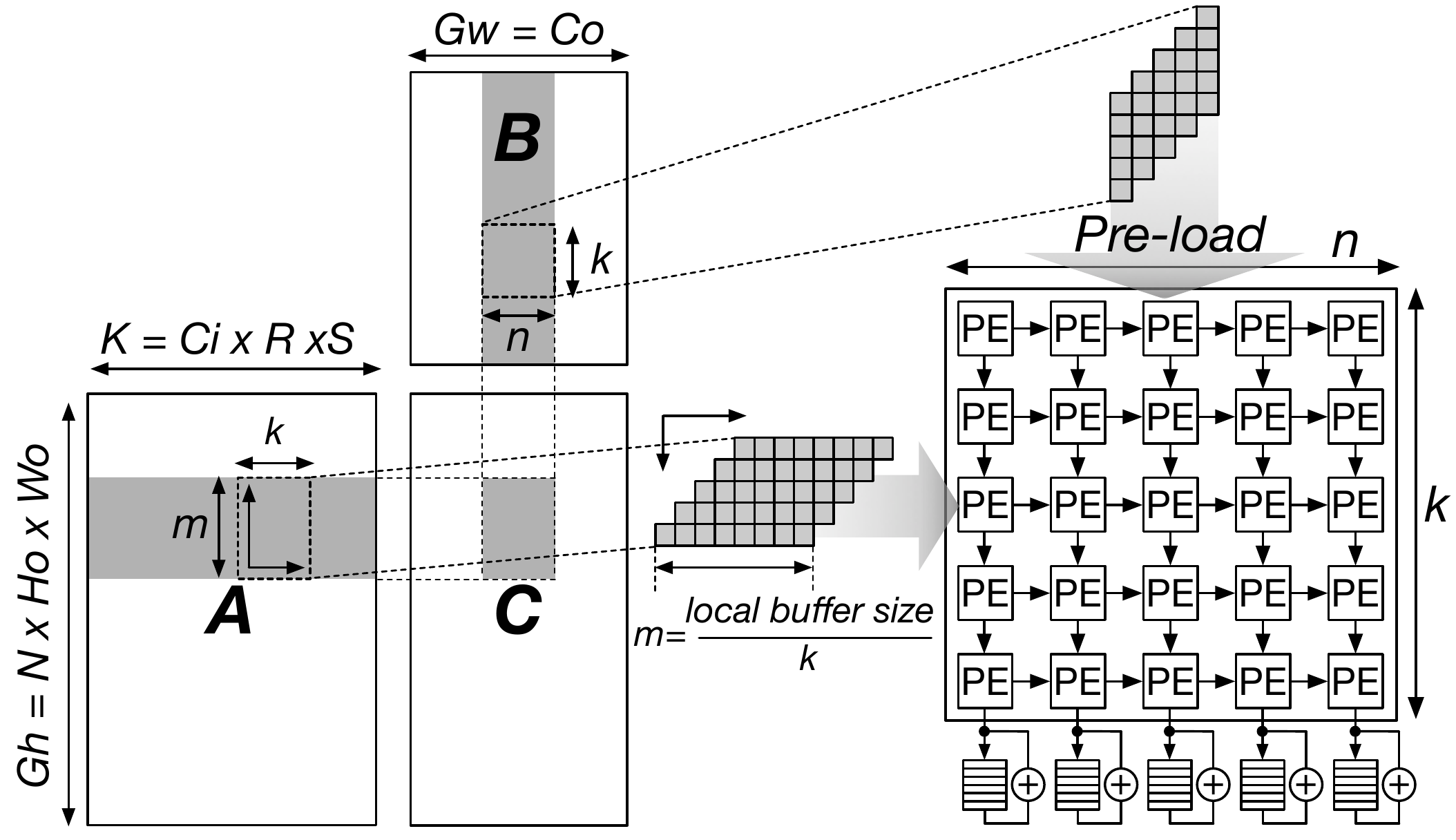}
    \vspace*{-2mm}
    \caption{GEMM dimensions , tiling, and mapping of each tile to the systolic array of a convolution layer in forward propagation.}
    \label{fig:im2col}
  \vspace*{-3mm}
\end{figure}

\begin{table}[t]
    \centering
    \caption{GEMM matrix dimensions for \emph{im2col} convolution for different CNN training phases. \(N, C, H, W\) indicate the sub-batch size, the channel count, and the height/width of each feature (\emph{i} and \emph{o} denote input and output features, respectively), and \(R, S\) are the height and width of each filter.}
    \label{tab:im2col}
    \vspace*{1mm}
    \noindent\resizebox{1.01\linewidth}{!}{
        \tabulinesep=0.5mm
        \renewcommand{\arraystretch}{0.7}
        \begin{tabu}{|[1.2pt]l|l|l|l|[1.2pt]}
            \thickhline
            Convolution Phase & \(G_h\) dimension & \(G_w\) dimension & K dimension \tabularnewline
            \midhline
            Forward & \(\color{red!70!black}{N \times H_o \times W_o}\) & \(C_o\) & \(C_i \times R \times S\) \tabularnewline
            \midhline
            Data Gradient & \(\color{red!70!black}{N \times H_i \times W_i}\) & \(C_i\) & \(C_o \times R \times S\) \tabularnewline 
            \hline
            Weight Gradient & \(C_i \times R \times S\) & \(C_o\) & \(\color{red!70!black}{N \times H_o \times W_o}\) \tabularnewline  
            \thickhline
        \end{tabu}
    }
    \vspace*{-4mm}
\end{table}

If the size of \(G_h\), \(G_w\), or \(K\) is smaller than the systolic array size, the compute units are significantly underutilized. However, with the networks we evaluate, this does not become a significant limitation: early layers with small sub-batches have large features and while the feature sizes of later convolution layers are small, their large sub-batch size compensates (shown by the red-colored dimensions in \tab{tab:im2col}).

\textbf{Systolic Dataflow for im2col GEMM.}
We block this im2col GEMM into multiple \(m \!\times\! n\) tiles, which are processed in sequence through the array. Each tile corresponds to a portion of the output matrix (C). The width of each tile is equal to the width of the systolic array (\(n\)). The height (\(m\)) is chosen to maximize the size of a tile, thus minimizing the number of tiles per layer and improving utilization: \(m=\frac{Local\ buffer\ size}{k=systolic\ array\ height}\). This is illustrated in \fig{fig:im2col}.

Each tile is processed using multiple waves through the systolic array, where each wave multiplies a block of input matrix A by a block of input matrix B. A block from B is first read one row at a time. Each row is shifted down until the array has one element of B per PE (this takes 5 cycles in the toy example of \fig{fig:im2col}). Then, a block of A is pipelined into the array with results for each element of C eventually accumulated at the bottom of the array as shown in the right side of \fig{fig:im2col}. Notice that in the figure, cycle 6 corresponds to the first row of the block of A having been multiplied and then accumulated by the first column of the block of B. In the following cycle, the second row of the A block completes its pipeline through the first column, while the first row of A now completes its dot product with the second column of the B block (and its output is at the bottom of the second column of the systolic array). 

Once a wave as described above completes, the next blocks of A and B are processed. As additional blocks are processed, their outputs are added to the current values of the C tile (a reduction across waves), eventually completing a tile of C in \(\ceil{K\!/k}\) waves (K is the dimension of the input matrices and k is the PE array height).

\begin{figure}[t]
    \centering
    \includegraphics[width=0.47\textwidth]{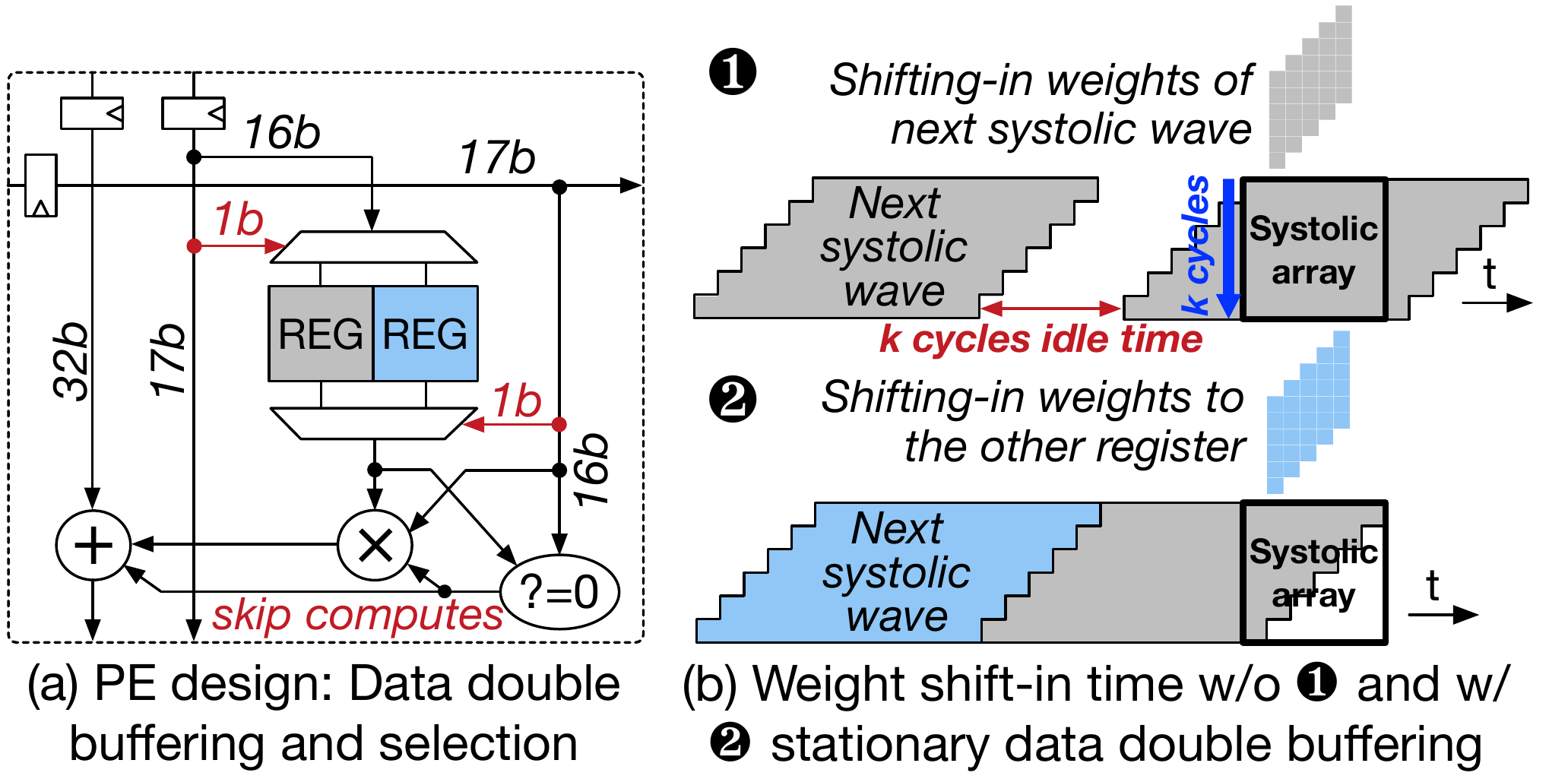}
    \vspace*{-2mm}
    \caption{Removing inter-wave idle time by weight double buffering and control signal shift.}
    \label{fig:arch_opt}
  \vspace*{-5mm}
\end{figure}

\textit{\textbf{Gap-less Waves with Weight Double Buffering.}}
The flow described above has one significant problem. Before every multiplication of blocks of A and B, the B block is read and distributed to the PEs, which requires k (PE array height) cycles (for reading and inter-PE shifting). No arithmetic occurs during these k cycles, which decreases performance (upper half of \fig{fig:arch_opt}{b}). 

To remove this inter-wave idle time, we modify the basic PE design to double buffer weights (\fig{fig:arch_opt}{a})---the next wave's weights are fetched and distributed into a second register within each PE while the current wave is still being processed. As the current wave starts draining from the PE array, the following wave starts immediately by feeding in a new block of A and multiplying by the second register that stores the next set of weights from B. Thus, there are no gaps between waves and an entire tile of C is computed without any idle time beyond the initial fill and final drain of the pipeline.
In addition to the extra register in each PE, a minor further change is that a select signal for choosing which weight register to use is propagated along with the inputs of A and B. This optimization significantly boosts performance at very low cost: the simple 1b local signal between every two PEs and a 16b register and multiplexer between the two registers per PE. As in prior work, we also check for zero inputs and skip arithmetic in such cases to reduce energy consumption~\cite{parashar2017scnn}.


\subsection{Overall Processor Architecture}
\label{sub_sec:proc_arch}
\begin{figure}
    \centering
    \includegraphics[width=0.42\textwidth]{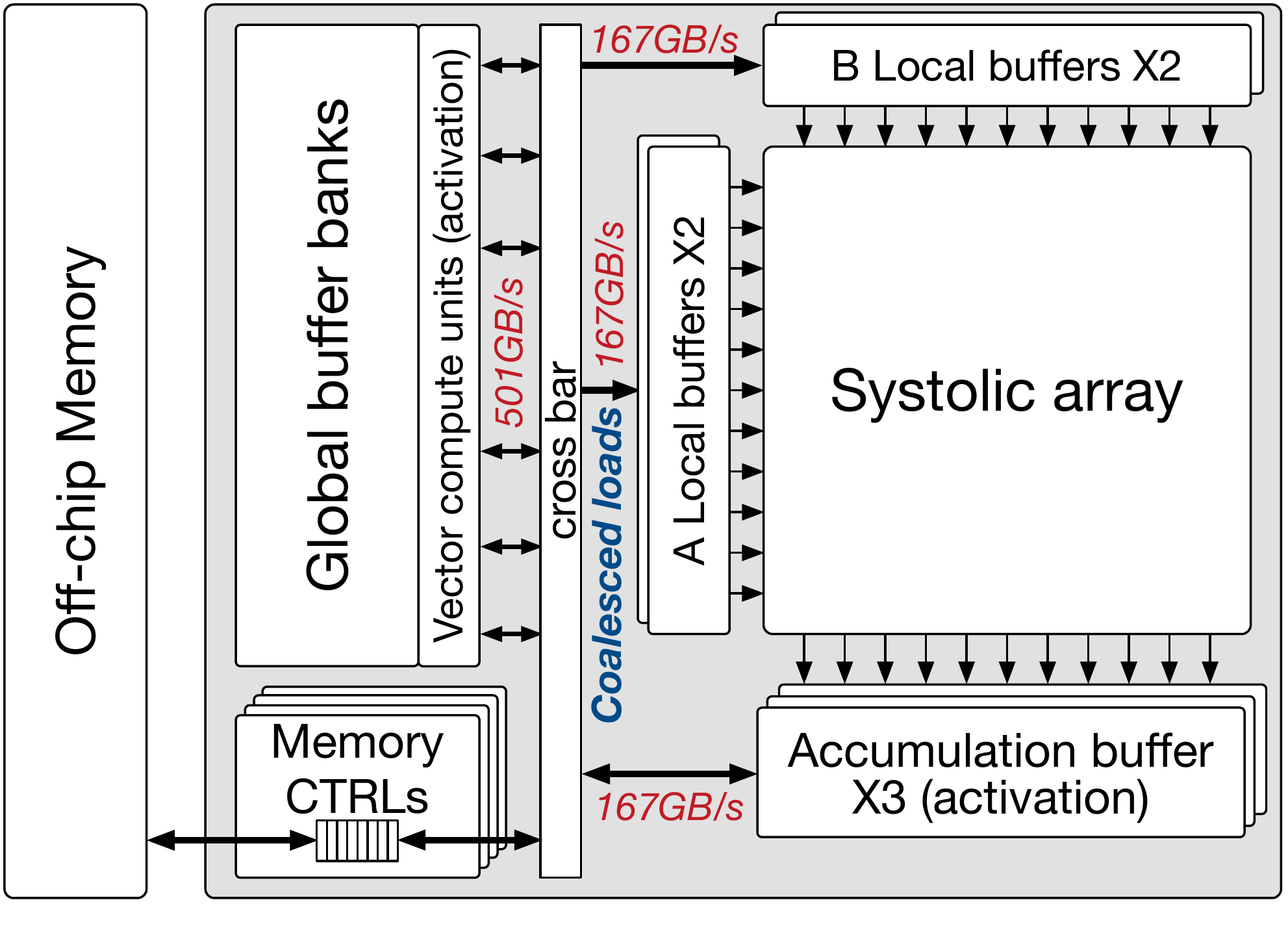}
    \vspace*{-4mm}
    \caption{Per-core architecture of the WaveCore accelerator.}
    \label{fig:arch}
  \vspace*{-4mm}
\end{figure}

In addition to the systolic cores, the WaveCore CNN training accelerator contains several more structures and units. \fig{fig:arch} illustrates the overall architecture of one core of the processor. There are two such cores in our  proposed design that are connected by an on-chip network, similar to TPU v2~\cite{tpuv2}. We describe these structures and estimate the area and power requirements of WaveCore below.

\textbf{Local Buffers.}
Both A and B local input buffers are double-buffered. Double buffering enables the overlap of computation within the PEs with accesses to the global buffer and to memory and allows for very simple coarse-grain control of data transfers between buffers and memory. We choose the minimal size for each buffer, such that PEs never directly access the global buffer or memory, as this avoids access-related stalls. A half-buffer of B stores a 16b word for each PE and is thus 32KiB (\(128 \!\times\! 128 \!\times\! 16b\)). Each half-buffer for A is 64KiB because A blocks need to be twice as large as B blocks to avoid inter-wave idle time. The output accumulation buffer is triple-buffered because it holds the current output tile while the previous tile is being written to memory and the partial gradient sums for the next tile are read. Each part of this buffer holds an entire tile of C and is 128KiB. Note that while outputs are summed in 32b precision, the final write to the output buffer quantizes to 16b precision.

\textbf{Global Buffers.}
The baseline global buffer is 10MiB and has 32 banks. This is sufficient for using \mbs with modern CNNs and avoiding bank access conflicts.
The global buffer is connected to all local buffers via a crossbar.
To avoid duplicated data loads from the global buffer, we have memory load coalescing units that maintain high effective bus bandwidth utilization. Our processor operates at a 0.7GHz clock frequency, and the data bandwidth of local and global buffers are set to fully support the systolic wave pipelining.

\begin{table}[b]
    \vspace*{-5mm}
    \caption{Accelerator specification and comparison.}
    \centering
\begin{threeparttable}
    \noindent\resizebox{1.0\linewidth}{!}{
        \tabulinesep=0.5mm
        \renewcommand{\arraystretch}{0.7}
        \begin{tabu}{|[1.2pt]l|[1.0pt]l|l|l|l|[1.2pt]}
            \thickhline
               & V100  & TPU v1   & TPU v2 & WaveCore  \tabularnewline
            \midhline
            Technology \((nm)\)         & 12 FFN          & 28   & N/A   & 32     \tabularnewline
            \hline
            Die Area \((mm^2)\)           & 812  & \(\leq 331\) & N/A & 534.0 \tabularnewline              
            \hline
            Clock Freq \((GHz)\)         & 1.53          & 0.7   & 0.7   & 0.7     \tabularnewline
            \hline
            TOPS / Die            & 125 (FP16)     & 92 (INT8) & 45 (FP16) & 45 (FP16)     \tabularnewline 
            \hline
            Peak Power \((W)\)        & 250         & 43     & N/A   & 56  \tabularnewline  
            \hline
            On-chip buffers \((MiB)\)        & 33\tnote{1}         & 24     & N/A   & 20 (2$\times$10)  \tabularnewline  
            \thickhline
        \end{tabu}
    }
         \begin{tablenotes}
         \scriptsize{\item[1] Sum of L2, shared memory, and registers}
         \end{tablenotes}
    \end{threeparttable}
        \label{tab:area}
    \vspace*{-5mm}
\end{table}

\textbf{Main Memory.}
The off-chip memory is connected to memory controllers, which communicate with the on-chip buffers via the crossbar switches. Our baseline WaveCore uses a single HBM2 stack with 4 dice~\cite{jedec2016hbm2}, which provides 8GiB off-chip DRAM with 300GiB/s data bandwidth over 8 channels (4 channels per core). We choose HBM2 because it is used by other modern training accelerators~\cite{tpuv2,volta2017whitepaper}. We later show that cheaper GDDR or even LPDDR memory can be sufficient for WaveCore.

\textbf{Vector and Scalar Computing Units.}
The systolic array is used for convolutions and fully-connected matrix operations, but cannot be efficiently utilized by normalization, pooling, and activation layers, which require a relatively small number of arithmetic operations. Such layers are memory bandwidth bound, and we therefore process them using scalar and simple vector units that are placed close to the global buffer where their outputs are then stored.

\textbf{Scalability.}
We describe and evaluate WaveCore with two cores, but compute throughput can be easily scaled with larger mini-batches distributed across multiple accelerators or additional cores.
As each accelerator or core conducts the same job, we can use \mbs within each WaveCore and only communicate for loss computation and parameter reduction and update.

\textbf{Area Estimation.}
We estimate the die area of WaveCore at 45nm technology and scale this estimate to 32nm to compare with other deep learning accelerators~\tab{tab:area}.  
The estimated total area of the two-core WaveCore is 534.0 \(mm^2\).
We use a 24T flipflop design as reported in~\cite{kim201427} and the floating point multiplier and adder designs reported in~\cite{hickmann2007parallel}.
Each PE requires 12,173 \(um^2\) and both multiplier and adder take more than 90\% of the PE area.
The estimated area of the 128\(\times\)128 PE array is 199.45 \(mm^2\), which accounts for 67\% of WaveCore's area.
The size of the global buffer and the vector compute units per core are estimated at 18.65 \(mm^2\) and 4.33 \(mm^2\), respectively.
The crossbar has 24 256b-wide ports (32B memory access granularity).
The area occupied by the network and the crossbar expands the chip width by 0.4mm, following the approach used to evaluate Dadiannao~\cite{chen2014dadiannao}.

\textit{\textbf{Power Modeling.}}
We use a convolution layer that exhibits 100\% systolic-array utilization to estimate the peak power consumption of WaveCore.
WaveCore operates at 0.7GHz, which is \(<\!1/2\) compared to V100 (1.53 GHz) and the same as TPU v2~\cite{tpuv2}.
WaveCore consumes a maximum of 56\(W\) (\tab{tab:area}).
Here, we use a HBM2 as the off-chip memory and model its power using the Rambus power model~\cite{vogelsang2010understanding} in 22nm technology.
The SRAM buffer power is calculated with CACTI~\cite{chen2012cacti} configured for 32nm.
The power consumed by multipliers and adders is taken from~\cite{han2016eie} and flipflops from~\cite{fuketa2013minimizing}.
The link and router power is calculated with Orion2.0~\cite{kahng2009orion}.


\section{Evaluation Methodology}
\label{sec:eval_method}
\begin{table}[t]
\vspace*{-3mm}
\caption{Evaluation configuration description.}
    \centering
    \noindent\resizebox{1.0\linewidth}{!}{
        \tabulinesep=0.5mm
        \renewcommand{\arraystretch}{0.7}
        \begin{tabu}{|[1.2pt]l|l|[1.2pt]}
            \thickhline
            \makecell[c]{Configuration}             & \makecell[c]{Description} \tabularnewline
            \midhline
            Baseline            & 2-level GEMM blocking \tabularnewline
            \hline
            ArchOpt             & Baseline + weight double buffering \tabularnewline
            \hline
            IL                 & ArchOpt + inter-layer data reuse \tabularnewline
            \hline
            {\mbs}-FS           & IL + serialize all layers using the same sub-batch size \tabularnewline
            \hline
            {\mbs}1             & IL +  greedy layer grouping \tabularnewline
            \hline
            {\mbs}2             & {\mbs}1 + inter-branch  data reuse \tabularnewline
            \thickhline
        \end{tabu}
    }
    \label{tab:config}
    \vspace*{-3mm}
\end{table}

\begin{table}[t]
    \vspace*{-2mm}
    \caption{Off-chip memory configuration.}
    \centering
    \noindent\resizebox{1.0\linewidth}{!}{
        \tabulinesep=0.5mm
        \renewcommand{\arraystretch}{0.7}
        \begin{tabu}{|[1.2pt]l|l|l|l|[1.2pt]}
            \thickhline
            \makecell[c]{Memory type}             & \makecell[c]{Per-chip configuration} & Chip \# & Total BW \tabularnewline
            \midhline
            HBM2       & \multirow{2}{*}{300 GiB/s, 8 GiB, 8 channels}  & x1 & 300 GiB/s \tabularnewline
            \cline{1-1} \cline{3-4}
            HBM2$\times$2       &   & x2 & 600 GiB/s \tabularnewline
            \hline
            GDDR5      & 32 GiB/s, 1GiB, 1 channel  & x12 & 384 GiB/s \tabularnewline
            \hline
            LPDDR4     & 29.9 GiB/s, 2GiB, 1 channel & x8 & 239.2 GiB/s \tabularnewline
            \thickhline
        \end{tabu}
    }
    \label{tab:mem}
    \vspace*{-3mm}
\end{table}

\begin{figure*}[!h]
    \centering
    \includegraphics[width=1.0\textwidth]{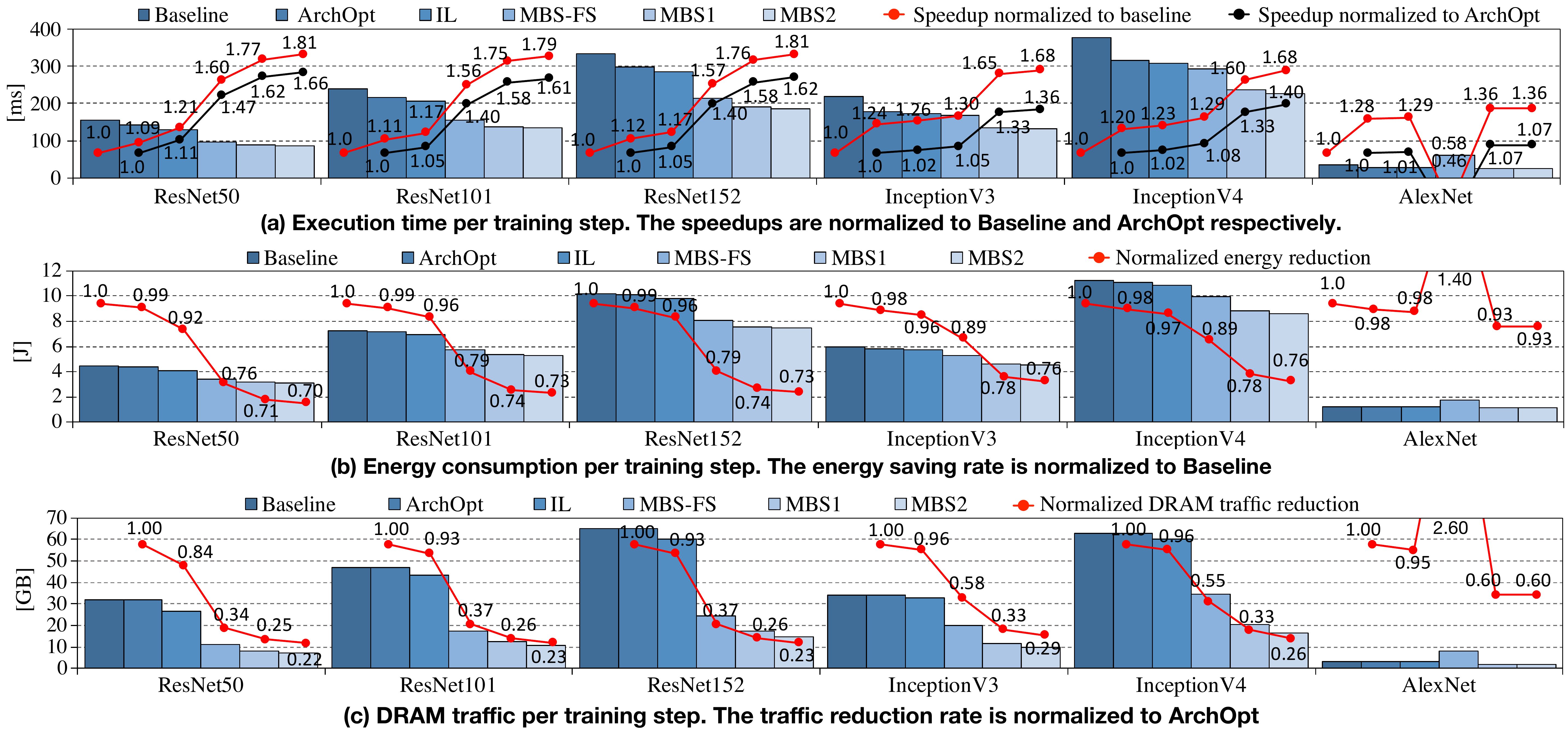}
    \vspace*{-7mm}
    \caption{DRAM traffic, performance and energy consumption sensitivity to the proposed network architecture reconfiguration and HW architecture optimization methods.}
    \label{fig:total}
  \vspace*{-3mm}
\end{figure*}

We evaluate the locality benefits of \mbs and the performance and energy of WaveCore on three well-known modern \emph{deep CNNs}: ResNet~\cite{he2016deep}, Inception v3~\cite{szegedy2015going}, and Inception v4~\cite{szegedy2017inception}. We also evaluate a shallower CNN (AlexNet~\cite{krizhevsky2012imagenet}) with few memory BW bound layers such as normalization and pooling. We use mini-batches of 32 samples per core (64 per chip) for the deep CNNs and 64 samples per core for AlexNet because of its smaller training context. We use 16b floating point for all CNNs with mixed-precision arithmetic (16b multiplication and 32b accumulation)~\cite{micikevicius2017mixed}.

For each network, we evaluate several execution configurations as summarized in \tab{tab:config}: \textbf{Baseline} uses two-level GEMM input matrix blocking for effective data reuse within each convolution and FC layer~\cite{kurzak2012autotuning}; \textbf{ArchOpt} adds weight double buffering for better PE utilization (all other configurations use ArchOpt), \textbf{Inter-Layer (IL)} reuses the shared data between layers but only when the per-layer memory footprint of the entire mini-batch fits within the on-chip buffer (i.e., not using the \mbs approach), \textbf{MBS-FS} is naive \mbs that fully serializes a mini-batch such that all layers in the CNN have the same sub-batch size, \textbf{MBS1} greedily forms layer groups to simultaneously optimize both intra- and inter-layer data reuse, and \textbf{MBS2} additionally reuses the inter-branch data which requires different layer grouping than MBS1. We compare WaveCore with \mbs to an NVIDIA TESLA V100 running Caffe~\cite{jia2014caffe} and report values averaged over 10 iterations.

The WaveCore simulator accounts for all memory, buffers, and on-chip interconnect traffic as well as the arithmetic operations. The default WaveCore uses a single HBM2 chip with 4Hi stacks. We also scale memory bandwidth using two HBM2 chips to launch a larger mini-batch per accelerator (and to more closely match commercial accelerators). Because \mbs significantly reduces memory traffic, we also evaluate lower-bandwidth main memory options that are cheaper and offer higher capacity (GDDR5 and LPDDR4). The off-chip memory configurations of WaveCore are listed in \tab{tab:mem}.


\section{Evaluation Results}
\label{sec:eval}

\fig{fig:total} compares the per-training-step execution time, energy consumption, and DRAM traffic of our proposed technique. In each of the subfigures, bars show absolute values and lines show relative ones. We normalize execution time separately to both Baseline and ArchOpt to isolate the impact of the architectural and algorithmic contributions of WaveCore and \mbs. 

Compared to Baseline, ArchOpt improves performance by 9--28\% across CNNs by removing the idle time between systolic waves. The gain is particularly large for AlexNet because AlexNet has mostly convolution layers with few memory-BW bound layers. Similarly, while not shown in the figure, ArchOpt provides more benefit with \mbs because the large reduction in memory traffic increases the relative impact of idle compute time. ArchOpt has little energy benefit ($\sim2\%$) as it conserves only static energy. 

Inter-layer (IL), which is similar to prior locality approaches used for inference, has only a modest impact on performance, energy, and traffic because many layers have large footprints that exceed the buffer size. 

MBS-FS, which uses a single sub-batch size (and thus a single group) substantially reduces DRAM traffic (42--66\%) for the deep CNNs because it utilizes inter-layer locality well. However, with a small sub-batch size, the time needed for the extra reads and writes of weight gradients used to accumulate them across sub-batches cannot be hidden, which reduces performance. This is evident in the performance trends of Inception v3 and v4, where MBS-FS is worse than IL. AlexNet exhibits a much larger performance loss with MBS-FS because it has three FC layers with large weights and the extra weight reads increase memory traffic by 2.6$\times$.

MBS1 balances inter- and intra-layer reuse and achieves large improvements in performance (33--62\%) and DRAM traffic (67--75\%) for the deep CNN compared to ArchOpt. AlexNet shows smaller gains as it lacks memory-BW bound layers. MBS1 also shows 22--29\% energy saving for the deep CNNs compared to Baseline by reducing the DRAM energy portion from 21.6\% to 8.7\%.
As WaveCore skips multiplication and addition when one of the inputs to a PE is zero, the contribution of DRAM traffic reduction to the overall energy saving is high.
It is important to note that global buffer traffic is increased by a similar amount as DRAM traffic is decreased.
However, there is still a large net energy saving because a global buffer access energy is $8\times$ lower than that of DRAM.

MBS2 reduces DRAM traffic by an additional 4--10\% and improves training performance by up to 7\% compared to MBS1.
MBS2 needs additional global buffer space to store the data at the shared multi-branch nodes, so the number of sub-batch iterations is larger than with MBS1. 
While more iterations imply a larger overhead for re-reading weights and gradients, the traffic saved by the reuse between branches is greater.
The gain is slightly bigger for Inception networks because the Inception modules have more branches and reuse opportunity scales with the number of branches.

In summary, the highly-optimized MBS2 improves DRAM traffic by 71--78\%, training performance by 36--66\%, and energy consumption by 24--30\% for the deep CNNs.

\begin{figure}[h]
\vspace*{-1mm}
    \centering
    \includegraphics[width=0.48\textwidth]{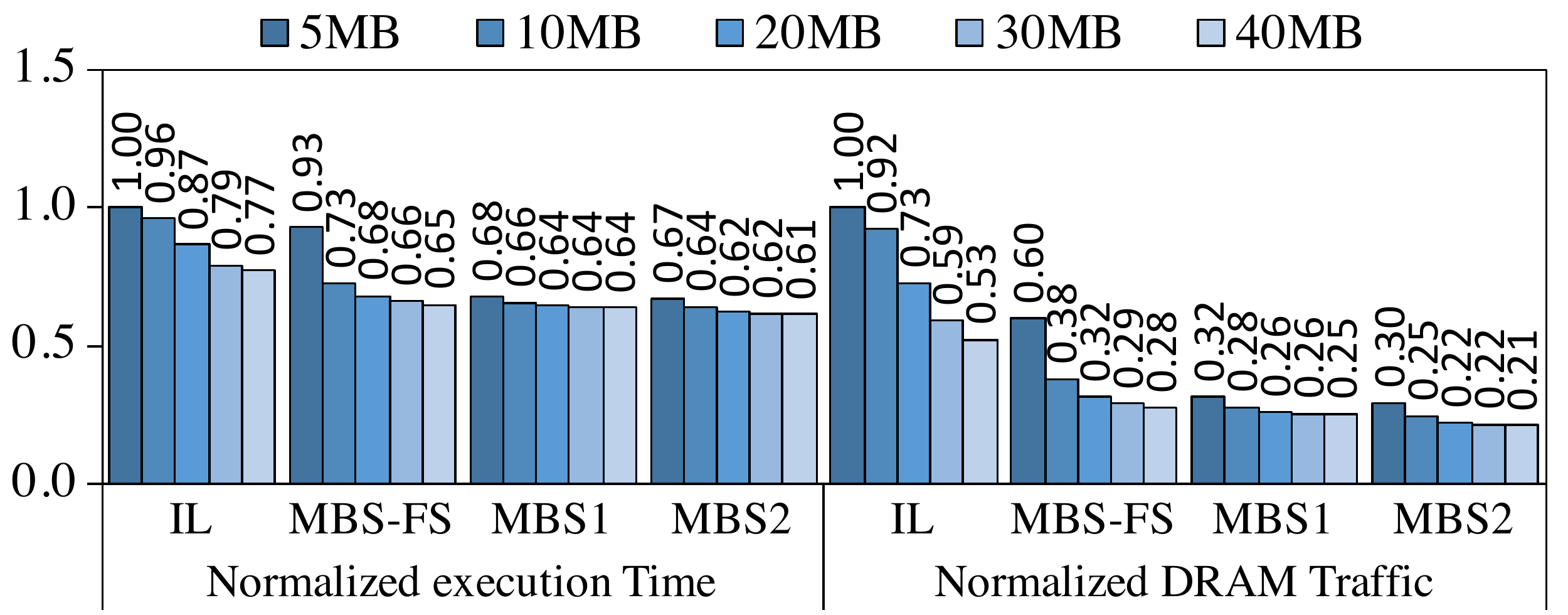}
    \vspace*{-7mm}
    \caption{Memory traffic and performance sensitivity of ResNet50 to the global buffer size (Normalized to IL with 5MiB).}
    \label{fig:gbuf_sensitivity}
\end{figure}

\textit{\textbf{Sensitivity to Global Buffer Size.}}
Another benefit of MBS is its low sensitivity to on-chip storage capacity. 
To showcase this, we compare the execution time and DRAM traffic per training step of ResNet50 for different configurations with different global buffer sizes (\fig{fig:gbuf_sensitivity}). 
The per-core global buffer size is scaled from 5MiB to 40MiB and execution time and traffic are normalized to IL at 5MiB (ResNet's \mbs scheduling requirement is smaller than 5MiB).
Even with a 40MiB global buffer, only 47\% of DRAM traffic is saved by IL; MBS2 saves 1.5X the traffic even with a 5MiB buffer.
IL with 40MiB also provides less performance benefit than both MBS1 and MBS2 at just 5MiB.
Both MBS1 and MBS2 show little performance and DRAM traffic variation for different buffer sizes because they simultaneously balance both intra- and inter-layer reuse.
In contrast to the optimized MBS1 and MBS2, MBS-FS again suffers from the impact of reading and writing gradient partial sums.

\begin{figure}[t]
\vspace*{-1mm}
    \centering
    \includegraphics[width=0.48\textwidth]{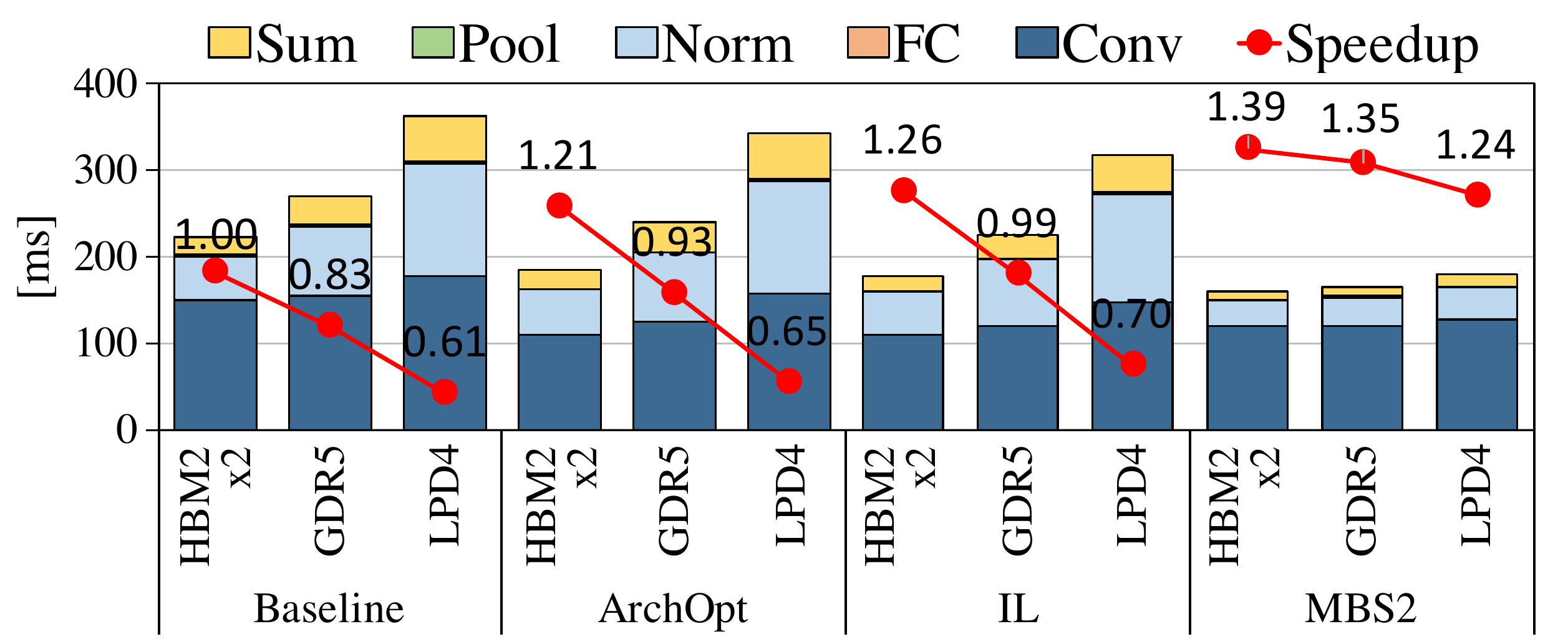}
    \vspace*{-6mm}
    \caption{ResNet50 training performance sensitivity to the memory type and the execution time breakdown by layer type.}
    \label{fig:mem_sensitivity}
  \vspace*{-3mm}
\end{figure}

\textit{\textbf{Sensitivity to DRAM BW.}}
\fig{fig:mem_sensitivity} highlights the ability of MBS to enable high performance even with lower-cost, lower-bandwidth memories. 
The figure compares the per-step training time of different configurations using various memory types (speedup is normalized to Baseline with 2$\times$HBM2).
The bandwidth of GDDR5 and LPDDR4 is 64\% and 40\% that of HBM2$\times2$, respectively.
While all implementations suffer from decreased bandwidth, the improved locality with MBS2 makes it far less sensitive with only a 4\% performance drop when using off-package GDDR5 and a $<\!15\%$ drop with low-cost LPDDR4. 
In this experiment, the off-chip memory space has been increased to 16GB to train 64 samples per core (128 per WaveCore) because off-package memories offer higher capacity.
\mattancut{Text below is fine, but seems unnecessary at this point of the paper.}{
ArchOpt shows the highest performance sensitivity to memory bandwidth (56\%) because it has significantly larger DRAM traffic than \mbs and has less compute unit idle time compared to Baseline. 
IL has similar performance sensitivity to ArchOpt because it has small DRAM traffic reduction.
MBS2, on the other hand, shows only 15\% performance penalty due to minor execution time increase in memory bandwidth bound layers potentially enabling to use a larger and cheaper off-chip memory system. 
}

\begin{figure}[h]
\vspace*{-1mm}
    \centering
    \includegraphics[width=0.47\textwidth]{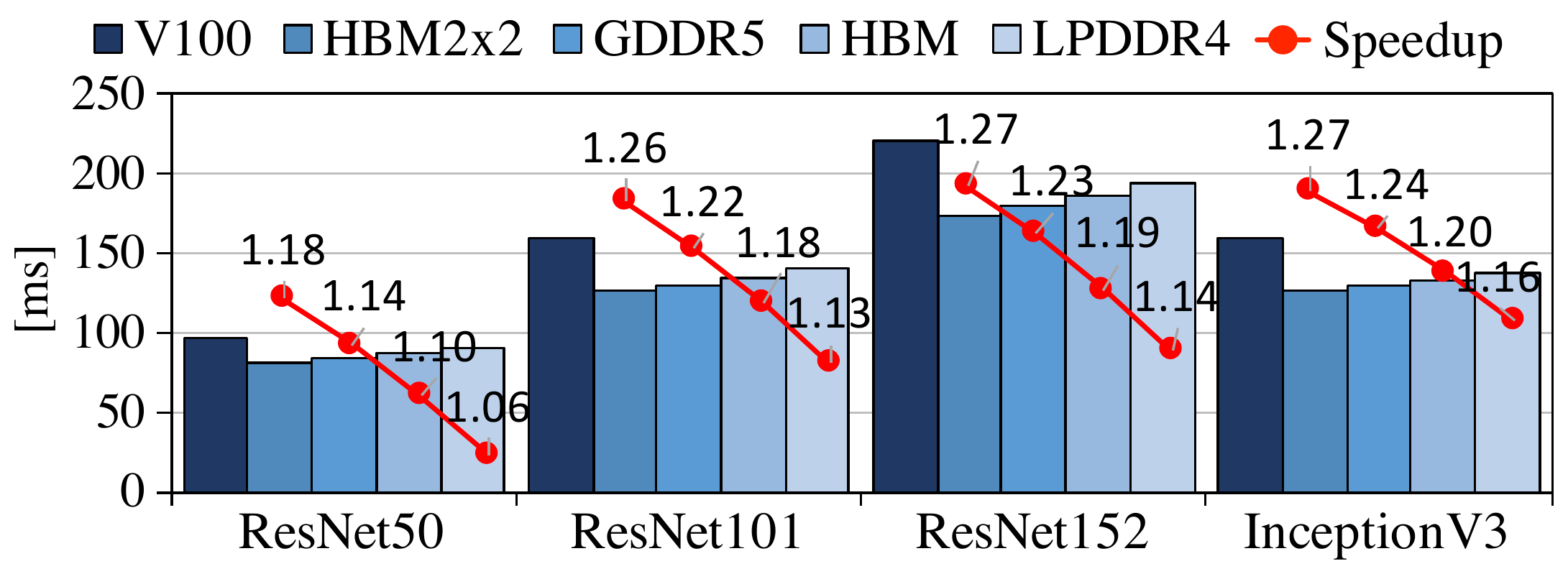}
    \vspace*{-3mm}
    \caption{NVIDIA V100 GPU performance comparison to WaveCore + MBS2 with different memory types.}
    \label{fig:gv100}
  \vspace*{-2mm}
\end{figure}

\textit{\textbf{Comparison to GPU.}}
\fig{fig:gv100} compares the measured execution time per training step of an NVIDIA V100 GPU with our estimates for WaveCore with different DRAM configurations.
Although a single WaveCore has 30\% the peak compute and 27\% the memory bandwidth (LPDDR4) of V100, it still exhibits better training performance.
The performance gap widens as the network depth increases because many layers with low data parallelism cannot efficiently utilize the wide compute resources of the  V100.
\mattancut{Agree with Armand, this is not necessary.}{Due to WaveCore's high compute efficiency, it is desirable to use multiple in tandem to form a highly efficient training environment similar to Google's TPU.}

\begin{figure}[t]
\vspace*{-1mm}
    \centering
    \includegraphics[width=0.48\textwidth]{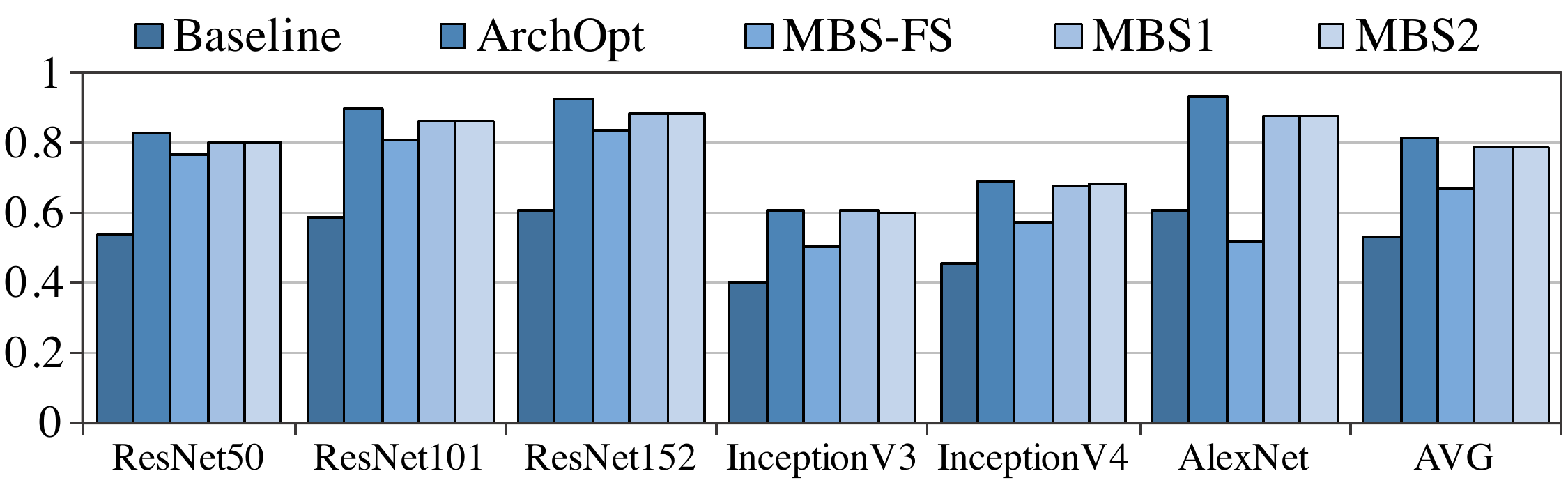}
    \vspace*{-8mm}
    \caption{Systolic array utilization of different CNNs.}
    \label{fig:systolic_util}
  \vspace*{-4mm}
\end{figure}
\textit{\textbf{Systolic Array Utilization.}}
As MBS propagates only a fraction of the mini-batch for each sub-batch iteration, it is important to observe its impact on the systolic core utilization.
\fig{fig:systolic_util} compares the utilization of convolution and FC layers for different CNNs.
To isolate the impact of sub-batch size and the parallelism it makes available on utilization, this experiment uses unlimited DRAM bandwidth.
Baseline suffers from low core utilization (average of 53.8\%) due to the inter-wave idle time. Double buffering with ArchOpt increases the average utilization to 81.5\%. 
MBS-FS exhibits lower utilization (66.7\%) because the sub-batch size is determined solely by the large early layers. 
Optimizing reuse with different sub-batch sizes across layer groups with MBS1 and MBS2 regains the lost utilization and brings it up to 78.6\%, within 3\% of a full mini-batch. 
This small difference is largely a result of a few early layers with small channel counts, which result in particularly narrow tiles that do not fully utilize WaveCore's 128\(\times \)128 systolic array. Later layers exhibit almost 100\% utilization.


\section{Related work}
\label{sub_sec:relatedwork}

To our knowledge, no prior work has addressed locality-optimizations for CNN training. Instead, we discuss methods proposed for inference accelerators.
Most inference accelerators optimize CNN scheduling to better utilize intra-layer locality~\cite{gao2017tetris,chen2017eyeriss,lu2017flexflow,chen2014diannao,du2015shidiannao,jouppi2017datacenter}.
They mainly focus on the data flow within a processing array, reducing data re-fetches by unrolling, or optimizing data access patterns within a convolution layer. 

SCNN~\cite{parashar2017scnn} is a scheduling method and architecture that reuses inter-layer data in CNN inference.
SCNN uses the on-chip buffer to hold both the input and output features of each layer along with all weights.
This is possible with a reasonable on-chip buffer size because SCNN relies on the fact that inference uses a single sample (mini-batch of 1), that features between layers are sparse because of ReLU, and that weights are even more sparse because they are pruned once training is complete. Together, an entire network can fit within an on-chip buffer.

However, the SCNN approach cannot be used for training because the same conditions do not hold true: mini-batches are large resulting in layer outputs that exceed buffer size, convolution layer outputs are not sparse, and weights are not sparse before pruning~\cite{han2015deep}. 

Fused-Layer CNN~\cite{alwani2016fused} is an inference flow that also utilizes inter-layer data.
The approach is to divide the initial input to the CNN (the input feature map) into tiles and propagate one tile through multiple layers.
Each convolution layer uses its input tile to produce a smaller output tile (because output cannot be produced for bands along the tile edges). 
The overlap between tiles is exploited via dedicated caches. While effective for the networks evaluated in~\cite{alwani2016fused}, Fused-Layer CNN can not be applied to training modern deep CNNs, because:
(1) convolution layers with small feature maps and large channel counts and weight data (deeper layers in modern CNNs) do not exhibit sufficient inter-tile locality;
(2) normalization layers are incompatible with the tiling used for the depth-first propagation;
(3) the inter-layer communication pattern in multi-branch modules, as well as in back propagation, is not only a direct communication between one layer to its following one;
and 
(4) tiles shrink as they are propagated depth-first through the network, which limits available parallelism and likely hurts PE utilization.

\section{Conclusion}
\label{sec:conclusion}

We introduce MBS, a mechanism to reuse the inter-layer data in CNN training and balance its locality with that of intra-layer data. MBS reconfigures the CNN computation graph by partitioning a mini-batch of samples into sub-batches whose memory footprint fits within on-chip storage. We show that MBS reduces the volume of DRAM accesses by up to 74\% while providing a high processing-element utilization of 79\%. Additionally, we are the first to demonstrate and exploit data reuse opportunities between branches in CNN multi-branch Residual and Inception modules. To efficiently use MBS CNN training, we introduce WaveCore, a systolic-array based CNN training accelerator. We design WaveCore to double-buffer data within its processing elements to remove idle time between the systolic waves used to compute the convolution and fully-connected layer outputs. Our evaluation demonstrates that we expect single WaveCore with MBS to achieve higher performance than one V100 GPU despite having the GPU having $3\times$ higher peak performance and memory bandwidth.

Furthermore, the high locality MBS achieved by balancing intra- and inter-layer reuse makes WaveCore very robust to memory design decisions. We demonstrate that both on-chip buffer capacity and available off-chip bandwidth have far smaller impact than using a conventional training approach. For example, even with a low-cost LPDDR4 DRAM system (the same DRAM used for mobile phones), WaveCore can outperform a high-end V100 GPU.


\section{acknowledgment}

The authors acknowledge Texas Advanced Computing Center (TACC) for providing HPC resources.

\bibliography{main.bbl}
\bibliographystyle{sysml2019}


\end{document}